\documentclass{article}

\usepackage{microtype}
\usepackage{graphicx}
\usepackage{subcaption}
\usepackage{hyperref}
\usepackage[preprint]{paper}
\usepackage{enumitem}
\usepackage{amsmath}
\usepackage{amssymb}
\usepackage{mathtools}
\usepackage{amsthm}
\usepackage[capitalize,noabbrev]{cleveref}
\theoremstyle{plain}

\theoremstyle{definition}

\theoremstyle{remark}

\usepackage[textsize=tiny]{todonotes}
\usepackage{multirow} 
\usepackage{threeparttable}
\usepackage{adjustbox}
\usepackage{booktabs}
\usepackage[table,xcdraw]{xcolor}

\icmltitlerunning{Quantize the Target, Quantize the Drafter: Efficient Inference with Qwen3.5-4B}
\begin{document}

\twocolumn[
  \icmltitle{Quantize the Target, Quantize the Drafter: Efficient Inference with Qwen3.5-4B}

  \icmlsetsymbol{equal}{*}    
  \begin{icmlauthorlist}
    \icmlauthor{Jaeyeon Kim}{equal,nota}
    \icmlauthor{Jewon Lee}{equal,nota}
    \icmlauthor{Bo-Kyeong Kim}{equal,nota}
  \end{icmlauthorlist}
  \icmlaffiliation{nota}{Nota Inc}
  \icmlcorrespondingauthor{Bo-Kyeong Kim}{bokyeong.kim@nota.ai}

  \vskip 0.3in
]

\printAffiliationsAndNotice{\icmlEqualContribution}

\begin{abstract}
This report describes our approach to the Efficient Qwen Competition, where the goal is to enable low-latency serving of Qwen3.5-4B on a resource-constrained NVIDIA A10G GPU. Our system combines a quantized target model with speculative decoding. To recover accuracy, we apply quantization-aware distillation to the target model while retaining the original quantization grid. To speed up decoding, a block-diffusion drafter specialized for the quantized target model is trained using a two-stage procedure: first learning from the high-precision target and then adapting to the low-precision target. Because the drafter is invoked at every speculative decoding step, we further reduce its overhead with quantization and sliding-window attention, preserving draft-token acceptance while improving long-context decoding latency. As a result, our submission achieves a 6.978$\times$ average speedup over the baseline while satisfying the required quality thresholds, ranking 3rd overall. We hope these results provide useful insights for practical LLM inference. The code and resources are available at \url{https://github.com/nota-github/adaptfm-quant-dflash}.
\end{abstract}

\section{Competition Overview}

The Efficient Qwen Competition~\cite{efficient-qwen-competition} focused on minimizing the inference latency of the Qwen3.5-4B~\cite{qwen3.5} large language model (LLM) on an AWS g5.xlarge instance equipped with a single NVIDIA A10G GPU (24 GB VRAM). Submissions were ranked by average speedup over the unoptimized baseline model across short, medium, and long input settings. A broad range of optimization techniques was permitted, provided that the resulting model satisfied the required quality thresholds on standard benchmarks.

\section{Our Approach}

This technical report describes our submission to the competition (see Table~\ref{tab:our_subm}), in which we ranked 3rd among more than 40 teams. Our approach combines quantization of Qwen3.5-4B with speculative decoding to accelerate inference (Figure~\ref{fig:approach_overview}):

\begin{figure}[t]
  \centering
    \includegraphics[width=0.88\linewidth]{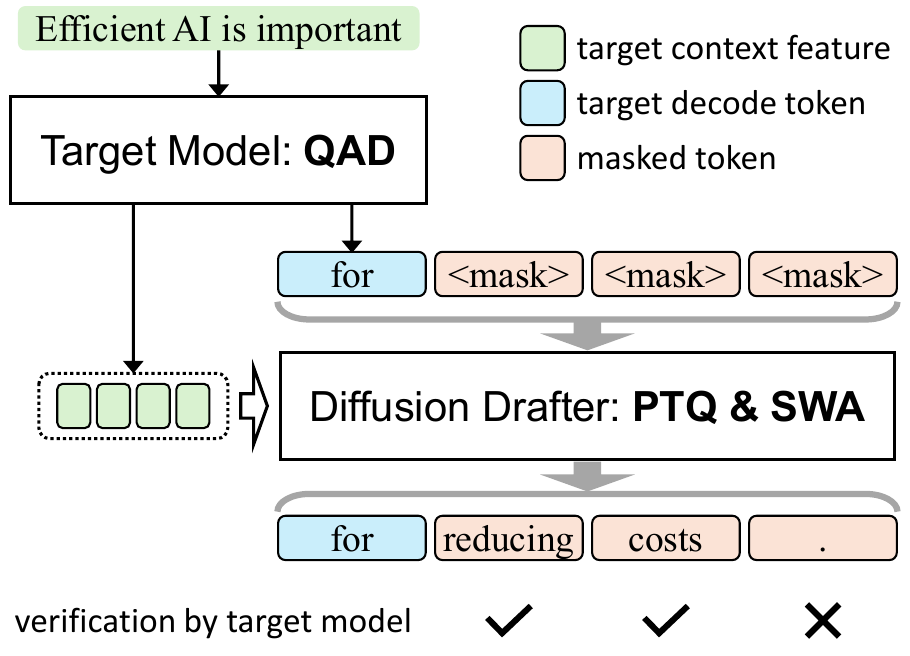}
  \vspace{-0.1em}  
  \caption{\textbf{Overview of our approach.} We use speculative decoding with a quantized target model and a quantized diffusion drafter equipped with sliding-window attention.}
  \label{fig:approach_overview}
\end{figure}

\begin{table}[t]
\centering

\begin{adjustbox}{max width=0.78\linewidth}
\begin{threeparttable}
\begin{tabular}{l|c|ccc|ccc}

\specialrule{.2em}{.1em}{.1em} 

Metric                & Baseline & Ours   \\ \hline\hline
Avg. Speedup          & 1.000×   & 6.978× \\ \hline
Short (64/128)\textsuperscript{a}        & 2582ms   & 232ms  \\
Medium (2048/256)\textsuperscript{a}     & 5441ms   & 782ms  \\
Long (8192/256)\textsuperscript{a}       & 6576ms   & 2313ms \\ \hline
MMLU-Pro ($\geq 0.621$)\textsuperscript{b}     & 0.690    & 0.659  \\
IFEval ($\geq 0.814$)\textsuperscript{b}       & 0.857    & 0.845  \\
GPQA-Diamond ($\geq 0.630$)\textsuperscript{b} & 0.700    & 0.667  \\ 
  
\specialrule{.2em}{.1em}{.1em} 

\end{tabular}

\begin{tablenotes}[para,flushleft]
\footnotesize 
\textsuperscript{a}Input/output token counts.
\textsuperscript{b}Required quality thresholds.

\end{tablenotes}
\end{threeparttable}

\end{adjustbox}

\vspace{0.1in}

\caption{\textbf{Our leaderboard entry.} We achieve a 6.978$\times$ average speedup over the unoptimized BF16 baseline on the NVIDIA A10G evaluation setting while satisfying the required quality thresholds (Team AFM-k5984d3s).}
\label{tab:our_subm}

\vspace{-0.15in}
\end{table}

\vspace{-0.7em}

\begin{enumerate}[itemsep=0em, leftmargin=1.2em]

\item[$\circ$] For the target model, we started from an existing post-training-quantized (PTQ) model checkpoint and applied quantization-aware distillation (QAD)~\cite{xin2026quantization} to recover accuracy.

\item[$\circ$] For speculative decoding, we trained a lightweight block-diffusion drafter~\cite{chen2026dflash} to operate with the quantized target model. We further optimized the drafter with PTQ and sliding-window attention (SWA)~\cite{beltagy2020longformer} to boost inference speed.

\end{enumerate}

\vspace{-0.7em}

\begin{figure}[t]
  \centering
    \includegraphics[width=\linewidth]{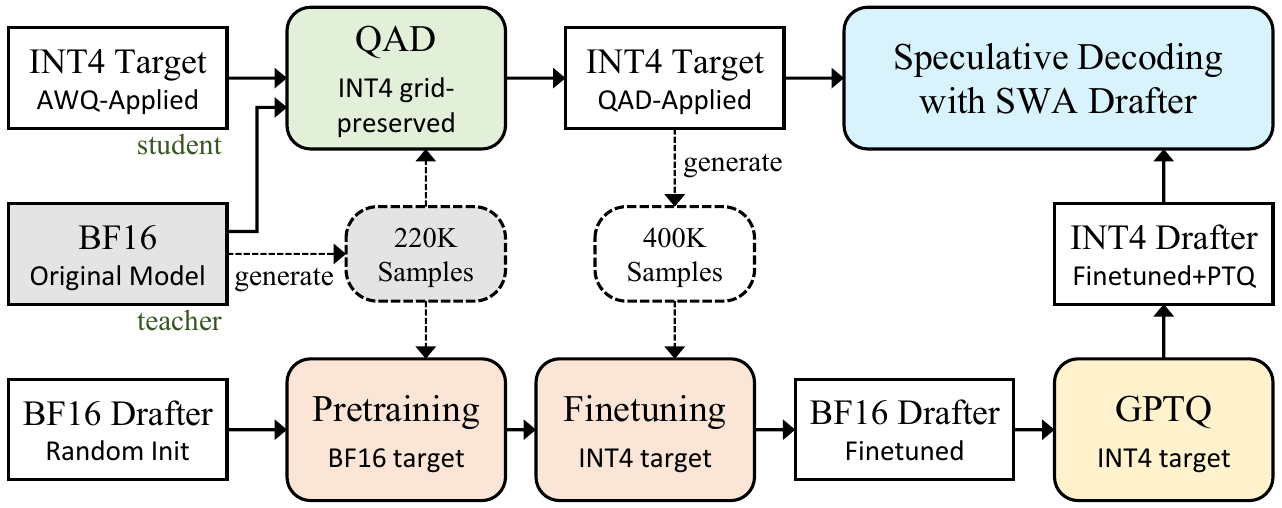}
    \caption{\textbf{Development pipeline.} QAD produces an INT4 target model while preserving the quantization grid. The block-diffusion drafter is trained in two stages: first with the BF16 target and then with the INT4 target. After finetuning, the drafter is quantized and equipped with SWA for speculative decoding.} \label{fig:dev_pipe}
\end{figure}

Figure~\ref{fig:dev_pipe} illustrates the training and optimization workflow used to build our final system. It highlights the sequence from target-model QAD to two-stage drafter training, followed by drafter PTQ and SWA for speculative decoding. Table~\ref{tab:main_overall_latency} summarizes the cumulative contribution of each component. We describe each technique in detail below.

\subsection{Target Model Optimization with QAD}

PTQ converts a pretrained model into a low-bit version without additional training, but it can suffer from accuracy degradation. To mitigate this, QAD~\cite{mishra2017apprentice,xin2026quantization} fine-tunes a quantized student model under the guidance of a high-precision teacher to recover accuracy.

We apply QAD to an AWQ~\cite{lin2024awq} INT4 checkpoint~\cite{qwen3.5-4b-awq-4bit} of Qwen3.5-4B. Specifically, we initialize the student from the dequantized model and optimize its weights while freezing the symmetric per-group scales. Using the original full-precision model as the teacher, we apply per-token distillation~\cite{hinton2015distilling} by minimizing the forward KL divergence: \(\mathbb{E}_{x}\left[\sum_t
\mathrm{KL}\left(p_{\text{tea}}(\cdot\mid x_{<t})
\,\|\,p_{\text{stu}}(\cdot\mid x_{<t}; Q_s(W))\right)\right]\),
where \(p_{\text{tea}}\) and \(p_{\text{stu}}\) denote the teacher and student output probabilities on a training sequence \(x\), respectively. Here, \(W_q\) denotes the INT4 codes with the frozen scales \(s\), and \(W\) is initially set to the dequantized weights \(s \odot W_q\). We update \(W\) through the fixed-grid fake quantizer \(Q_s(\cdot)\) using a straight-through estimator~\cite{bengio2013estimating}.

At packing time, we round the updated BF16 weights back onto the same quantization grid, yielding new INT4 codes under the original AWQ scales. The resulting model is served in the W4A16 compressed-tensors format.

\subsection{Speculative Decoding with DFlash}

Speculative decoding~\cite{xia2024unlocking} speeds up generation by using a fast draft model to propose several tokens, which are then verified in parallel by the target model. DFlash~\cite{chen2026dflash} further improves drafting efficiency with a diffusion-based block drafter that leverages the target model's hidden states to predict multiple future tokens in a single forward pass. 

For faster LLM inference, we train the DFlash drafter on top of the frozen QAD-applied target model. Given an input sequence, hidden representations are extracted from multiple layers of the target model and projected into compact context features. The target context is then injected into the drafter's key-value cache for conditioning. The drafter is trained to predict the masked tokens in each draft block, where an observed anchor token is followed by masked future positions. This encourages the drafter to propose future tokens aligned with the quantized target model's predictions.

We use a two-stage training pipeline for the drafter. First, we train the drafter from a random initialization using the original BF16 target model and data regenerated by this model. We then fine-tune the Stage 1 drafter with the QAD-applied target model using data regenerated by the QAD target. This two-stage design aims to learn a robust drafting initialization from the BF16 model before adapting it to the quantized target distribution. 

\begin{table}[t]
\centering

\begin{adjustbox}{max width=\linewidth}
\begin{threeparttable}
\begin{tabular}{l|c|ccc}

\specialrule{.2em}{.1em}{.1em} 

Method                                                                           & \begin{tabular}[c]{@{}c@{}}Ave.\\ Speedup\end{tabular} & \begin{tabular}[c]{@{}c@{}}Short\\ (64/128)\textsuperscript{$\dagger$}\end{tabular} & \begin{tabular}[c]{@{}c@{}}Medium\\ (2048/256)\textsuperscript{$\dagger$}\end{tabular} & \begin{tabular}[c]{@{}c@{}}Long\\ (8192/256)\textsuperscript{$\dagger$}\end{tabular} \\ \hline \hline
BF16 Baseline                                                                    & 1.00×                                                  & \begin{tabular}[c]{@{}c@{}}2487ms\\ (1.00×)\end{tabular} & \begin{tabular}[c]{@{}c@{}}5167ms\\ (1.00×)\end{tabular}    & \begin{tabular}[c]{@{}c@{}}5692ms\\ (1.00×)\end{tabular}  \\ \hline
\begin{tabular}[c]{@{}l@{}}+ INT4 target model\\ \ \  (AWQ→QAD)\end{tabular}      & 2.16×                                                  & \begin{tabular}[c]{@{}c@{}}1102ms\\ (2.26×)\end{tabular} & \begin{tabular}[c]{@{}c@{}}2308ms\\ (2.24×)\end{tabular}    & \begin{tabular}[c]{@{}c@{}}2876ms\\ (1.98×)\end{tabular}  \\ \hline
\begin{tabular}[c]{@{}l@{}}+ BF16 drafter\\ \ \  (Two-step training)\end{tabular} & 3.32×                                                  & \begin{tabular}[c]{@{}c@{}}642ms\\ (3.87×)\end{tabular}  & \begin{tabular}[c]{@{}c@{}}1350ms\\ (3.83×)\end{tabular}    & \begin{tabular}[c]{@{}c@{}}2522ms\\ (2.26×)\end{tabular}  \\ \hline
\begin{tabular}[c]{@{}l@{}}+ INT4 drafter\\ \ \  (GPTQ)\end{tabular}              & 3.55×                                                  & \begin{tabular}[c]{@{}c@{}}625ms\\ (3.98×)\end{tabular}  & \begin{tabular}[c]{@{}c@{}}1232ms\\ (4.19×)\end{tabular}    & \begin{tabular}[c]{@{}c@{}}2290ms\\ (2.49×)\end{tabular}  \\ \hline
+ Drafter with SWA                                                               & 3.57×                                                  & \begin{tabular}[c]{@{}c@{}}624ms\\ (3.98×)\end{tabular}  & \begin{tabular}[c]{@{}c@{}}1318ms\\ (3.92×)\end{tabular}    & \begin{tabular}[c]{@{}c@{}}2027ms\\ (2.81×)\end{tabular}  \\ 
              
\specialrule{.2em}{.1em}{.1em} 

\end{tabular}

\begin{tablenotes}[para,flushleft]
\footnotesize 

\textsuperscript{$\dagger$}Input/output token counts.
\newline
\textbullet \ Evaluated using 5 warm-up samples and 50 measurement samples from LongBench v2.
\end{tablenotes}
\end{threeparttable}

\end{adjustbox}

\vspace{0.1in}
\caption{\textbf{Contribution of each optimization to Qwen3.5-4B inference speedup.} Latency is measured on an NVIDIA RTX 5000 Ada Generation GPU. Each row incrementally adds one optimization on top of the previous row.}\label{tab:main_overall_latency}

\vspace{-0.2in}
\end{table}

\subsection{Drafter Optimization with PTQ and SWA}

We optimize the trained drafter with PTQ to improve end-to-end latency. Since the drafter is invoked at every speculative decoding step, reducing its computation directly lowers the per-step cost. We show that PTQ does not noticeably degrade drafter quality and largely preserves the acceptance length of draft tokens, allowing the reduced per-step cost to translate into lower overall latency. Since several PTQ methods yield similar performance in our experiments, we adopt GPTQ~\cite{frantar2022gptq} for drafter quantization.

We also equip the PTQ drafter with SWA~\cite{beltagy2020longformer}. Unlike full-context attention, SWA restricts the drafter's attention to a local window of recent tokens, reducing attention computation. This also allows the drafter to avoid distant context that may act as noise in long-context decoding, helping it focus on more relevant context for predicting near-future tokens. In our experiments, SWA often increases the mean acceptance length and reduces decoding latency, especially for long-context inputs.

\section{Experimental Setup}

\subsection{Target-Model QAD}

We generate the training data by prompting the original Qwen3.5-4B with 220K samples randomly drawn from Nemotron-Post-Training-Dataset-v2~\cite{NemotronPostTrainingDatasetV2}. The resulting teacher-regenerated conversations are tokenized and packed into fixed-length, EOS-delimited sequences with a sequence length of 16,384, padding only the last chunk. This yields 27K training chunks, corresponding to 440M tokens. 

We initialize the student from the dequantized weights of the AWQ INT4 checkpoint and distill it from the frozen BF16 teacher using the per-token forward KL objective. During distillation, we update only the dequantized weights while keeping the per-group scales frozen. We optimize the student with 8-bit AdamW at a learning rate of 2e-6 using a cosine schedule with a warmup ratio of 0.03, gradient clipping of 1.0, and an effective batch size of 16. Training runs for 8,000 steps, corresponding to about 4.5 epochs. We select the 5,000-step checkpoint as it yields the highest average performance among periodically evaluated checkpoints on a validation subset comprising SciQ~\cite{SciQ-welbl2017crowdsourcing}, BoolQ~\cite{BoolQ-clark2019boolq}, PIQA~\cite{bisk2020piqa}, and HellaSwag~\cite{zellers2019hellaswag}.

\subsection{Drafter Training}

We train a 5-layer DFlash drafter with hidden size 2,560 and block size 16, resulting in approximately 537M BF16 parameters. We use a two-stage setup: Stage 1 trains the drafter from random initialization using the original BF16 target model and its 220K regenerated conversations, which are also used for target-model QAD. Stage 2 finetunes the Stage-1 drafter using the QAD-applied INT4 target model and 400K conversations regenerated from Nemotron-Post-Training-Dataset-v2. For Stage 2, the target weights are dequantized from the INT4 weight grid, keeping it numerically aligned with the deployed INT4 target while allowing floating-point computation.

Training runs for 2 epochs with an effective batch size of 32. We optimize the drafter with AdamW using a learning rate of 1e-3, a cosine-annealing schedule with a warmup ratio of 0.04, and gradient clipping of 1.0. We set the maximum sequence length to 3,072, the number of anchors to 512, and the loss-decay factor $\gamma$ to 7.0.

\subsection{Drafter Optimization}

\textbf{Drafter PTQ.} For drafter PTQ calibration, we use 256 samples from the same 400K training set generated by the target INT4 model. Unlike token-driven GPTQ calibration, the DFlash drafter is conditioned on target hidden states, noise embeddings, and position IDs rather than token IDs. Therefore, we do not truncate samples to a fixed sequence length. Instead, each sample is kept as a full teacher-forced conversation at its natural length. We accumulate the GPTQ Hessian from 16 strided anchor positions per conversation over the assistant region. At each anchor, the drafter receives the causal target-hidden prefix and one 16-token draft block, yielding 256 $\times$ 16 = 4,096 calibration blocks. The drafter weights are then quantized to group-wise symmetric INT4 with a group size of 128.

\textbf{Drafter SWA.} To reduce speculative-decoding latency on long contexts, we apply SWA only to the DFlash drafter, using a fixed-size window of either 1,024 or 2,048 tokens. Since the drafter performs bidirectional attention over mask-token queries, we use a symmetric window around each query rather than a causal left-only window. The window is applied only at the attention-kernel level while keeping the KV cache full, so no cache eviction or rotation is introduced. For shorter sequences within the window size, the drafter falls back to full attention.

\subsection{Evaluation}

\textbf{Accuracy.} Following the competition rules, we evaluate model quality on MMLU-Pro~\cite{wang2024mmlu} ($\geq 0.621$), IFEval~\cite{zhou2023instruction} ($\geq 0.814$), and GPQA-Diamond~\cite{rein2023gpqa} ($\geq 0.630$), where the values in parentheses denote the required quality thresholds.

\textbf{Speed.} We follow the evaluation format with three length regimes: Short (64/128 input/output tokens), Medium (2,048/256), and Long (8,192/256). Since the exact official prompts are not publicly available, we measure the mean acceptance length using samples from GSM8K~\cite{cobbe2021training}, HumanEval~\cite{chen2021evaluating}, and LongBench v2~\cite{bai2025longbench}, with 64 prompts per benchmark and generation capped at 256 new tokens. For end-to-end latency, we use 5 warm-up samples and 50 measurement samples from LongBench v2. Due to the unavailability of an NVIDIA A10G GPU, latency is evaluated on an NVIDIA RTX 5000 Ada Generation GPU.


\begin{table}[t]
\centering

\begin{adjustbox}{max width=\linewidth}
\begin{threeparttable}
\begin{tabular}{l|ccc}

\specialrule{.2em}{.1em}{.1em} 

Target Model          & \multicolumn{1}{c}{\begin{tabular}[c]{@{}c@{}}MMLU-Pro\\ ($\geq 0.621$)\end{tabular}} & \multicolumn{1}{c}{\begin{tabular}[c]{@{}c@{}}IFEval\\ ($\geq 0.814$)\end{tabular}} & \multicolumn{1}{c}{\begin{tabular}[c]{@{}c@{}}GPQA-Diamond\\ ($\geq 0.630$)\textsuperscript{$\dagger$}\end{tabular}} \\ \hline \hline
BF16 Baseline  & 0.6804                                                                          & 0.8249                                                                        & 0.6798 {\scriptsize $\pm 0.0210$ }                                                                     \\
INT4 PTQ       & 0.6557                                                                          & 0.8046                                                                        & 0.6536 {\scriptsize $\pm 0.0132$ }                                                                     \\
\rowcolor[HTML]{ECF4FF} 
INT4 PTQ + QAD & 0.6610                                                                          & 0.8285                                                                        & 0.6626 {\scriptsize $\pm 0.0207$ }                                                                     \\ 
   
\specialrule{.2em}{.1em}{.1em} 

\end{tabular}

\begin{tablenotes}[para,flushleft]
\footnotesize 
\textsuperscript{$\dagger$}GPQA-Diamond reports mean $\pm$ std over five runs.
\newline
\end{tablenotes}
\end{threeparttable}

\end{adjustbox}

\caption{\textbf{Effect of QAD on Qwen3.5-4B benchmark accuracy.} QAD improves the AWQ INT4 checkpoint~\cite{qwen3.5-4b-awq-4bit} while retaining the original quantization grid. Values in parentheses denote the competition quality gates.} \label{tab:qad_accuracy}

\vspace{-0.1in}
\end{table}

\begin{table}[t]
\centering

\begin{adjustbox}{max width=\linewidth}
\begin{threeparttable}

\begin{tabular}{lcc|c|ccc}

\specialrule{.2em}{.1em}{.1em} 

\begin{tabular}[c]{@{}l@{}}Drafter\\ Setting\end{tabular}       & \begin{tabular}[c]{@{}c@{}}Drafter\\ Init\end{tabular} & \begin{tabular}[c]{@{}c@{}}Target\\ Model\end{tabular} & Ave  & GSM8K & \begin{tabular}[c]{@{}c@{}}Human\\ Eval\end{tabular} & \begin{tabular}[c]{@{}c@{}}Long\\ Bench v2\end{tabular} \\ \hline \hline
\begin{tabular}[c]{@{}l@{}}Direct\\ training\end{tabular} & Random                                                 & INT4                                                   & 4.97 & 6.37  & 5.42                                                 & 3.10                                                    \\ \hline
\begin{tabular}[c]{@{}l@{}}Stage-1\\ pretraining\end{tabular}   & Random                                                 & BF16                                                   & 4.92 & 5.75  & 5.41                                                 & 3.59                                                    \\
\rowcolor[HTML]{ECF4FF} 
\begin{tabular}[c]{@{}l@{}}Stage-2\\ finetuning\end{tabular}    & Stage-1                                                & INT4                                                   & 5.03 & 6.28  & 5.53                                                 & 3.30                                                    \\ 

\specialrule{.2em}{.1em}{.1em} 

\end{tabular}

\begin{tablenotes}[para,flushleft]
\footnotesize 
\textbullet \ The public DFlash checkpoint~\cite{zlab2026qwen35dflash} with the BF16 target yields an average acceptance length of 5.69 across three benchmarks.
\newline
\textbullet \ Measured with 64 prompts per benchmark and a 256-token generation cap.
\end{tablenotes}
\end{threeparttable}

\end{adjustbox}

\vspace{0.1in}
\caption{\textbf{Mean acceptance length of trained DFlash drafters.}
Two-stage training slightly outperforms direct training on the quantized target model.} \label{tab:drafter_training}

\vspace{-0.2in}
\end{table}

\begin{table}[t]
\centering

\begin{adjustbox}{max width=\linewidth}
\begin{threeparttable}

\begin{tabular}{l|c|ccc}

\specialrule{.2em}{.1em}{.1em} 

Drafter Quant  & Ave  & GSM8K & HumanEval & LongBench v2 \\ \hline\hline
BF16 Finetuned & 5.03 & 6.28  & 5.53      & 3.30         \\ \hline
INT4 RTN       & 4.91 & 6.00  & 5.47      & 3.24         \\
INT4 AWQ       & 4.92 & 6.07  & 5.46      & 3.24         \\
\rowcolor[HTML]{ECF4FF} 
INT4 GPTQ      & 4.98 & 6.13  & 5.49      & 3.30         \\ 

\specialrule{.2em}{.1em}{.1em} 
\end{tabular}

\begin{tablenotes}[para,flushleft]
\footnotesize 
\textbullet \ Measured with 64 prompts per benchmark and a 256-token generation cap.
\end{tablenotes}
\end{threeparttable}

\end{adjustbox}

\vspace{0.1in}
\caption{\textbf{Effect of drafter quantization on mean acceptance length.} PTQ preserves drafter quality while improving decoding speed, particularly in the long-input setting, as demonstrated by the latency results in Table~\ref{tab:main_overall_latency}.} \label{tab:drafter_quant}

\vspace{-0.1in}
\end{table}

\begin{table}[t]
\centering

\begin{adjustbox}{max width=\linewidth}
\begin{threeparttable}

\begin{tabular}{l|ccc|ccc}

\specialrule{.2em}{.1em}{.1em} 

\multirow{2}{*}{\begin{tabular}[c]{@{}l@{}}Drafter SWA\\ Window Len\end{tabular}} & \multicolumn{3}{c|}{BF16-Target + BF16-Drafter} & \multicolumn{3}{c}{W4-Target + W4-Drafter} \\
                                                                                  & Short          & Medium         & Long          & Short        & Medium       & Long         \\ \hline \hline
Full attention                                                                    & 1165.1         & 2144.8         & 4285.7        & 625.3        & 1232.4       & 2290.0       \\ \hline

\rowcolor[HTML]{ECF4FF} 
SWA 2048                                                                          & 1164.7         & 2220.6         & 3218.8        & 624.6        & 1250.3       & 1966.0       \\

\rowcolor[HTML]{ECF4FF} 
SWA 1024                                                                          & 1165.2         & 2502.8         & 3653.3        & 624.2        & 1318.4       & 2026.5       \\
SWA 512                                                                           & 1165.6         & 2794.8         & 3972.0        & 623.5        & 1410.8       & 2136.3       \\
SWA 256                                                                           & 1138.7         & 3166.4         & 4365.7        & 623.7        & 1537.1       & 2233.1       \\ 

\specialrule{.2em}{.1em}{.1em} 
\end{tabular}

\begin{tablenotes}[para,flushleft]
\footnotesize 
\textbullet \ Evaluated using 5 warm-up samples and 50 measurement samples from LongBench v2.
\textbullet \ Input/output tokens: Short 64/128, Medium 2048/256, Long 8192/256.

\end{tablenotes}
\end{threeparttable}

\end{adjustbox}

\vspace{0.1in}
\caption{\textbf{Effect of SWA on end-to-end latency (ms).} Limiting the drafter attention window improves long-context decoding speed, especially with 1024--2048 window lengths. Latency is measured in milliseconds on an NVIDIA RTX 5000 Ada Generation GPU.} \label{tab:drafter_swa}

\vspace{-0.1in}
\end{table}

\section{Results}

\textbf{Target-model QAD.} Table~\ref{tab:qad_accuracy} shows that direct INT4 PTQ degrades accuracy compared with the BF16 baseline, most notably dropping IFEval below the required quality gate. Applying QAD improves the INT4 checkpoint across all three benchmarks, recovering IFEval from 0.8046 to 0.8285 and increasing MMLU-Pro and GPQA-Diamond to 0.6610 and 0.6626, respectively. As a result, the QAD model satisfies all competition quality gates while retaining the original AWQ quantization grid. Since the subsequent speculative decoding step is lossless, these accuracy results are preserved in the final optimized model.

\textbf{Drafter training.} Table~\ref{tab:drafter_training} compares direct drafter training on the INT4 target with our two-stage training pipeline. Since the drafter is conditioned on target hidden states and predicts masked future tokens in each draft block, its acceptance length depends on how well its proposals align with the target model. The Stage-1 drafter trained with the BF16 target achieves an average acceptance length of 4.92, showing that the BF16 target provides a useful initialization for block drafting. After adapting this drafter to the QAD-applied INT4 target in Stage 2, the average acceptance length increases to 5.03. This slightly outperforms direct training from random initialization on the INT4 target, which obtains 4.97 on average, with gains on HumanEval and LongBench v2. These results suggest that learning a robust drafting initialization from the BF16 target before adapting to the quantized target distribution is beneficial.

\textbf{Drafter PTQ.} Table~\ref{tab:drafter_quant} compares INT4 PTQ methods for the drafter, including the round-to-nearest (RTN) baseline, AWQ, and GPTQ. INT4 quantization causes only a minor change in the drafter's mean acceptance length relative to the BF16 baseline, and these PTQ methods perform comparably. We therefore use GPTQ for drafter quantization. Because the drafter runs at every speculative decoding step, reducing its cost with PTQ improves the overall decoding speed, as reflected in Table~\ref{tab:main_overall_latency}.

\textbf{Drafter SWA.} Table~\ref{tab:drafter_swa} shows that SWA is most effective in the long-input setting. With a 2048-token window, long-input latency decreases from 4285.7 ms to 3218.8 ms in the BF16 case, and from 2290.0 ms to 1966.0 ms in the W4 case. For the leaderboard entry, we use a 1024-token SWA window, which corresponds to the latency reported in Table~\ref{tab:main_overall_latency}. A moderate SWA window reduces unnecessary drafter attention cost while preserving enough context for stable draft-token acceptance. Smaller windows, such as 512 or 256 tokens, are less effective because the reduced context can lower acceptance length and offset the savings from cheaper attention. We therefore use a 1024- or 2048-token SWA window for the drafter.

\section{Conclusion}

This report presents an approach for accelerating Qwen3.5-4B inference for efficient deployment while maintaining model accuracy. Our system combines a low-bit target model improved by quantization-aware distillation with speculative decoding. We train a block-diffusion drafter specialized for the quantized target model and further optimize the drafter with quantization and sliding-window attention to reduce decoding overhead. Together, these optimizations achieve a 6.978× average speedup over the baseline on an NVIDIA A10G GPU.

\section*{Acknowledgments}

We thank the AI Industrial Convergence Cluster Development project funded by the Ministry of Science and ICT (MSIT, Korea) and Gwangju Metropolitan City for their generous support of GPU resources.

\bibliography{paper}
\bibliographystyle{paper}

\end{document}